# An Empirical Evaluation of Possible Variations of Lazy Propagation

**Anders L Madsen**
Hugin Expert A/S
Niels Jernes Vej 10
9220 Aalborg E, Denmark
Anders.L.Madsen@hugin.com

## Abstract

As real-world Bayesian networks continue to grow larger and more complex, it is important to investigate the possibilities for improving the performance of existing algorithms of probabilistic inference. Motivated by examples, we investigate the dependency of the performance of Lazy propagation on the message computation algorithm.

We show how Symbolic Probabilistic Inference (SPI) and Arc-Reversal (AR) can be used for computation of clique to clique messages in the addition to the traditional use of Variable Elimination (VE).

In addition, the paper presents the results of an empirical evaluation of the performance of Lazy propagation using VE, SPI, and AR as the message computation algorithm. The results of the empirical evaluation show that for most networks, the performance of inference did not depend on the choice of message computation algorithm, but for some randomly generated networks the choice had an impact on both space and time performance. In the cases where the choice had an impact, AR produced the best results.

## 1 Introduction

The Bayesian network (Pearl 1988; Cowell *et al.* 1999; Jensen 2001; Neapolitan 2004) formalism offers an intuitive and compact graphical model representation for reasoning under uncertainty. As Bayesian network models continue to grow larger and more complex, the efficiency of algorithms for (exact) probabilistic inference becomes more and more important. One task is to identify enhancements of existing inference algorithms, which may improve both the space and time performance of probabilistic inference.

Algorithms for probabilistic inference in Bayesian networks can be classified into two different classes. The first class of algorithms is referred to as direct computation (query-based) algorithms as they perform inference based on the structure of the graph of the Bayesian network. This class includes VE (Cannings *et al.* 1978; Zhang and Poole 1994), Bucket Elimination (Dechter 1996), SPI (Shachter *et al.* 1990; Li and D'Ambrosio 1994), AR (Olmsted 1983; Shachter 1986), the Fusion operator (Shenoy 1997), and Belief Propagation (Pearl 1988).

Using a query-based algorithm, the task of probabilistic inference is usually defined as computing $P(T \mid \epsilon)$ for a target set $T$ given evidence $\epsilon$. Prior to solving a query $Q$, the graph of the Bayesian network is pruned to remove any variables not relevant for $Q$. This pruning proceeds by $d$-separation analysis and removal of barren variables. This analysis is performed for each query. By performing this *non-local* operation prior to performing inference, it is possible to exploit the independence relations induced by $\epsilon$.

The second class of algorithms for probabilistic inference is referred to as indirect computation algorithms as they proceed by message passing in a secondary computational structure such as a junction tree (also known as join tree or cluster tree). This class includes Lauritzen-Spiegelhalter (Lauritzen and Spiegelhalter 1988), Hugin (Jensen *et al.* 1990), and Shenoy-Shafer propagation (Shenoy and Shafer 1990).

Using an indirect computation algorithm, the task of probabilistic inference is usually defined as computing $P(X \mid \epsilon)$ for each variable $X$ given evidence $\epsilon$. A junction tree structure is constructed from the graph of the Bayesian network. Usually, this structure is constructed once and *off-line*, but used to solve all subsequent inference tasks. This structure should be large enough to propagate all possible dependence relations of the original Bayesian network given any subset of evidence. These algorithms are truly local computation algorithms in the sense that no global analysis of the graph is performed.

Lazy Propagation (Madsen and Jensen 1999) is an inference algorithm, which combines direct and indirect computation for computing all posterior marginals often outperforming the traditional indirect computation algorithms. It performs message passing based on the scheme of Shenoy-Shafer propagation in a junction tree, but messages are computed by direct computation using a variable elimination approach.

Motivated by examples, we investigate the dependency of the performance of Lazy propagation on the direct computation algorithm used for message computation. We show how SPI and AR can be used for computation of clique to clique messages in addition to the traditional use of VE, which is equivalent to Bucket Elimination and the Fusion operation.

The results of the empirical evaluation show that for most networks, the performance of inference did not depend on the choice of message computation algorithm, but for some randomly generated networks the choice had an impact on both space and time performance. In the cases where the choice had an impact, AR produced the best results.

In summary, this paper contributes with results on both theoretical and empirical issues related to the use of AR and SPI for message computation in Lazy propagation.

## 2    Preliminaries and Notation

We assume the reader to be familiar with most concepts of basic graph theory. A mixed graph $G = (V, E)$ consists of a set of vertices $V$ and a set of edges $E \subseteq V \times V$ where $E$ may contain both directed and undirected edges. An edge $(X_i, X_j)$ is called undirected if both $(X_i, X_j) \in E$ and $(X_j, X_i) \in E$, whereas an edge $(X_i, X_j)$ is directed with head $X_j$ and tail $X_i$ if $(X_i, X_j) \in E$ and $(X_j, X_i) \notin E$. We only consider finite acyclic mixed graphs. The set of descendants $\text{de}(X)$ of $X$ in a directed graph $G$ is the set of vertices, which can be reached from $X$ by a directed path.

A discrete Bayesian network $\mathcal{N} = (G, \mathcal{P})$ consists of an acyclic, directed graph $G = (V, E)$ and a set of conditional probability distributions $\mathcal{P}$. The vertices $V$ of $G$ correspond one-to-one with the variables of $\mathcal{P}$. $\mathcal{N}$ induces a joint probability distribution over $V$:

$$P(V) = \prod_{X \in V} P(X \mid \text{pa}(X)), \qquad (1)$$

where $\text{pa}(X)$ is the set of variables corresponding to the parents of the vertex representing $X$ in $G$ and $\text{fa}(X) = \text{pa}(X) \cup \{X\}$.

Let $\phi(H \mid T)$ be a probability potential with head $H = \text{H}(\phi)$ and tail $T = \text{T}(\phi)$ (i.e., a non-negative function on $H \cup T$). The domain $\text{dom}(\phi)$ of $\phi$ is defined as $\text{dom}(\phi) = H \cup T$. The domain graph $G(\{\phi\})$ induced by $\phi$ is defined as $G = (\text{dom}(\phi), \{(H_1, H_2), (H_2, H_1) \mid H_1, H_2 \in \text{H}(\phi)\} \cup \{(T, H) \mid H \in \text{H}(\phi), T \in \text{T}(\phi)\})$. The domain graph of a set of potentials $\Phi$ is defined as $G(\Phi) = \bigcup_{\phi \in \Phi} G(\{\phi\})$. In general, a domain graph will be a mixed graph. An undirected edge is, for instance, created when a parent $Y$ of two non-adjacent variables $X_1$ and $X_2$ is eliminated, see Figure 1.

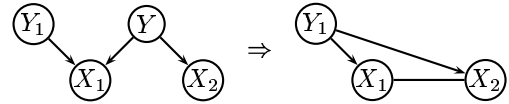

Figure 1: An undirected edge is introduced when $Y$ is eliminated.

The moral graph $G^m$ of $G$ is obtained by adding undirected edges between all pairs of vertices with a common child and all pairs with children connected by an undirected path and dropping direction of all directed edges.

## 3    Probabilistic Inference

We define the task of probabilistic inference as follows:

**Definition 3.1 [Probabilistic Inference]**
Given a Bayesian network model $\mathcal{N} = (G = (V, E), \mathcal{P})$ and a set of evidence $\epsilon$, probabilistic inference is the task of computing $P(X \mid \epsilon)$ for all $X \in V \setminus \epsilon$.

In this paper, we consider only hard evidence (i.e., an instantiation of variable $X$ to $X = x$). Soft evidence can be handled, but does not produce the same performance improvements as hard evidence. Conceptually, evidence $\epsilon = \{X = x\}$ is inserted into $\mathcal{N}$ by removing all outgoing arcs from $X$ and instantiating the distribution $P(X \mid \text{pa}(X))$ obtaining $P(X = x \mid \text{pa}(X))$ and the distribution of each $Y \in \text{ch}(X)$ to reflect $X = x$ where $\text{ch}(X)$ are the children of $X$ in $G$.

We define a query $Q$ to be a triple $Q = (\Phi, T, \epsilon)$ where $\Phi$ is a set of probability potentials (e.g., $\mathcal{P}$ of $\mathcal{N} = (G, \mathcal{P})$), $T$ is a set of variables (the target), and $\epsilon$ is the set of evidence.

**Definition 3.2 [Barren Variable]**
Let $Q = (\Phi, T, \epsilon)$ be a query. A variable $X$ is a *barren variable* w.r.t. $Q$, if $X \notin T$, $X \notin \epsilon$, and all descendants $\text{de}(X)$ of $X$ are barren.

This definition of barren variables can be extended to the case of mixed graphs. If $G(\Phi)$ is a mixed graph, then we consider each maximum set of vertices connected by undirected edges only as one vertex when determining barren variables w.r.t. a query $Q = (\Phi, T, \epsilon)$. The set of variables $W$ relevant for a query $Q = (\Phi, T, \epsilon)$ is the set of variables not separated (Cowell *et al.* 1999) from $T$ and not barren w.r.t. $Q$.

## 4　Potentials

We redefine the notion of a potential.

**Definition 4.1 [Potential]**
A *potential* on $W \subseteq V$ is a singleton $\pi_W = (\Phi)$ where $\Phi$ is a set of non-negative real functions on subsets of $W$.

The domain graph $G$ induced by a potential $\pi = (\Phi)$ is defined as the graph $G(\Phi)$ induced by $\Phi$ and denoted by $G(\pi)$. We call a potential $\pi_W$ *vacuous* if $\pi_W = (\emptyset)$.

We define the operations of combination and contraction as follows:

**Definition 4.2 [Combination]**
The *combination* of potentials $\pi_{W_1} = (\Phi_1)$ and $\pi_{W_2} = (\Phi_2)$ denotes the potential on $W_1 \cup W_2$ given by $\pi_{W_1} \otimes \pi_{W_2} = (\Phi_1 \cup \Phi_2)$.

**Definition 4.3 [Contraction]**
The contraction $\mathrm{c}(\pi_W)$ of a potential $\pi_W = (\Phi)$ is the non-negative function on $W$ given as $\mathrm{c}(\pi_W) = \prod_{\phi \in \Phi} \phi$.

Using the notion of potentials as defined above, we define a query as $Q = (\pi, T, \epsilon)$. The solution to a query $Q$ is $\pi^{\downarrow T} = (\{\phi_1, \ldots, \phi_n\})$ where the operation of marginalization is defined for each direct computation algorithm in the following subsections.

When the definition of marginalization has been established, it is easily shown that for $W_1 \subseteq W$ we have:

$$\mathrm{c}(\pi_W^{\downarrow W_1}) = \sum_{W \setminus W_1} \mathrm{c}(\pi_W).$$

## 5　Lazy Propagation

The basic idea of lazy propagation is to perform inference in a junction tree structure maintaining decompositions of clique and separator potentials until combination becomes mandatory by variable elimination. By construction a junction tree is wide enough to support the computation of any posterior marginal given evidence on any subset of variables. The junction tree is, however, often too wide to take advantage of independence properties induced by evidence. Lazy propagation is aimed at taking advantage of independence and irrelevance properties induced by evidence in a Shenoy-Shafer message passing scheme.

### 5.1　Initialization

The first step in initialization of the junction tree representation $\mathcal{T} = (\mathcal{C}, \mathcal{S})$ of $\mathcal{N}$ is to associate a vacuous potential with each clique $C \in \mathcal{C}$. Next, for each node $X$, we assign $P(X \mid \mathrm{pa}(X)) \in \mathcal{P}$ to a clique $C$, which can accommodate it, i.e., $\mathrm{fa}(X) \subseteq C$. It is not necessary to assign $P$ to the smallest such clique.

The initialization of $\mathcal{T}$ is completed once each $P(X \mid \mathrm{pa}(X)) \in \mathcal{P}$ has been associated with a clique of $\mathcal{T}$.

After initialization each clique $C$ holds a potential $\pi_C = (\Phi)$. The set of clique potentials is invariant during propagation of evidence. The joint potential $\pi_V$ on $\mathcal{T} = (\mathcal{C}, \mathcal{S})$ is:

$$\pi_V = \bigotimes_{C \in \mathcal{C}} \pi_C = \left( \bigcup_{X \in V} \{P(X \mid \mathrm{pa}(X))\} \right).$$

The contraction of the joint potential $\pi_V$ is equal to the joint probability distribution of $\mathcal{N}$:

$$\mathrm{c}(\pi_V) = \mathrm{c}(\bigotimes_{C \in \mathcal{C}} \pi_C) = \prod_{X \in V} P(X \mid \mathrm{pa}(X)).$$

Evidence is inserted after initialization.

### 5.2　Message Passing

Lazy propagation employs a Shenoy-Shafer message passing scheme. The message passing is controlled by preselecting a clique $R \in \mathcal{C}$ as the root of $\mathcal{T}$. Message passing in $\mathcal{T}$ is performed via the separators $\mathcal{S}$. The separator $S = A \cap B$ between two adjacent cliques $A$ and $B$ stores the messages passed between $A$ and $B$. First, messages are passed from the leaf cliques of $\mathcal{T}$ to the root $R$ by recursively letting each clique $A$ pass a message to its parent $B$ whenever $A$ has received a message from each of its children, see Fig. 2. Secondly, messages are passed in the opposite direction.

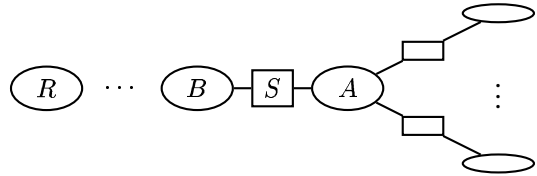

Figure 2: A junction tree with root clique $R$.

### 5.3　Messages

The message $\pi_{A \to B}$ is passed from clique $A$ to clique $B$ by absorption. Absorption from $A$ to $B$ involves eliminating the variables $A \setminus B$ from the combination of the potential associated with $A$ and the messages passed to $A$ from adjacent cliques $\mathrm{adj}(A)$ except $B$. In principle the message $\pi_{A \to B}$ is computed as:

$$\pi_{A \to B} = \left(\pi_A \otimes \left(\otimes_{C \in \mathrm{adj}(A) \setminus \{B\}} \pi_{C \to A}\right)\right)^{\downarrow B},$$

where $\pi_{C \to A}$ is the message passed from $C$ to $A$.

The algorithm for computing the message $\pi_{A \to B}$ is:

**Algorithm 5.1 [Compute Message]**
Let $A$ and $B$ be two adjacent cliques with separator $S = A \cap B$. The message $\pi_{A \to B}$ is computed as:

1. Let $\pi_{AB} = \pi_A \otimes (\otimes_{C \in \mathrm{adj}(A) \setminus \{B\}} \pi_{C \to A}) = (\Phi_{AB})$.

2. Let $\pi_{AB}^R = (\Phi_{AB}^R)$ where $\Phi_{AB}^R \subseteq \Phi_{AB}$ is the subset of potentials relevant for computing $\pi_{A \to B}$.

3. Solve $Q = (\pi_{AB}^R, S, \epsilon)$ by direct computation to obtain $\pi_{A \to B}$.

4. (Minimize the tail of each potential of $\pi_{A \to B}$).

5. Store the message $\pi_{A \to B}$ at $S$.

Based on the separation property in undirected graphs (Cowell *et al.* 1999), we use a breadth-first-search algorithm to identify the set of probability potentials $\Phi_{AB}^R$ relevant for a message $\pi_{A \to B}$. Starting with the set containing all potentials $\phi$ with $S \cap \mathrm{dom}(\phi) \neq \emptyset$, we iteratively include all potentials sharing a variable with a potential in the current set of potentials. Subsequently, we remove from $\Phi_{AB}$ all potentials with only barren head variables.

Motivated by the necessity of minimalization of Conditional Gaussian (CG) potential tails in order to be able to perform the initial propagation in the Lauritzen-Jensen architecture for inference in mixed Bayesian networks (Lauritzen and Jensen 2001), we consider minimalization of tails as a new optional step (step 4) of Lazy propagation. A potential tail is minimal if no tail variable is independent of all head variables.

## 6 Lazy VE Propagation

Lazy propagation (Madsen and Jensen 1999) using VE for message computation is here referred to as Lazy VE propagation.

**Algorithm 6.1 [VE Message Computation]**
To solve $Q = (\pi = (\Phi), T, \epsilon)$ by VE do:

1. Moralize $G = G(\pi)$ to obtain $G^m = (V, E)$.

2. Let $\sigma$ be an elimination order for $V \setminus T$.

3. Eliminate each variable $Y \in V \setminus T$ in order of $\sigma$

    (a) Set $\Phi_Y := \{\phi \in \Phi \mid Y \in \mathrm{dom}(\phi)\}$.

    (b) Eliminate $Y$:

    $$\phi_Y = \sum_Y \prod_{\phi \in \Phi_Y} \phi$$

    (c) Set $\Phi := \Phi \setminus \Phi_Y \cup \{\phi_Y\}$.

4. Return $\pi^{\downarrow T} = \pi(\Phi)$.

The correctness of Lazy VE propagation was established by (Madsen and Jensen 1999).

## 7 Lazy SPI Propagation

SPI is a direct computation algorithm, which is fundamentally different from VE. The idea of SPI is to solve a query as a combinatorial optimization problem (Li and D'Ambrosio 1994). Instead of focusing on the order in which variables are eliminated, SPI focuses on the order in which potentials should be combined. In this sense SPI is a more fine-grained algorithm than VE. SPI is similar to binary join trees (Shenoy 1997).

Since the basic idea of Lazy propagation is to maintain decompositions of clique and separator potentials until combination becomes mandatory by variable elimination, we need to make some adjustments to SPI before it is applicable for message computation.

Before solving a query $Q = (\pi, T, \epsilon)$, we can determine the number of potentials $\{\phi_1, \ldots, \phi_n\}$ in the solution $\pi^{\downarrow T}$ and the set of potentials $\Phi_i$ involved in the computation of each potential $\phi_i$. This implies that we can decompose $\Phi$ into disjoint subsets of non-idle potentials referred to as source potentials (the subset of potentials relevant for computing a potential $\phi_i$ is its set of source potentials). An idle potential is a potential not involved in any computation. Each set of source potentials produce one potential of $\pi_{A \to B} = \{\phi_1, \ldots, \phi_n\}$. This observation builds on the fact that the set of fill-in edges produced and therefore also the potentials created is independent of the order in which variables are eliminated (Rose *et al.* 1976).

**Algorithm 7.1 [SPI Message Computation]**
To solve $Q = (\pi = (\Phi), T, \epsilon)$ by SPI do:

1. Set $\Phi^I := \{\phi \in \Phi \mid \mathrm{dom}(\phi) \subseteq T\}$.

2. Set $\Phi := \Phi \setminus \Phi^I$.

3. Identify subsets $\Phi_1, \ldots, \Phi_n$ of source potentials.

4. For each set of source potentials $\Phi_i$

    (a) Initialize combination candidate set $B := \emptyset$.

    (b) Repeat

i. Add all pairwise combinations of elements of $\Phi_i$ to $B$ not already in $B$, except combinations of marginal factors without a common child.

ii. Select a pair $p = \{\phi_i, \phi_j\}$ of $B$ according to some criteria.

iii. If variables $W$ can be eliminated, then
- Set $\Phi_i := \Phi_i \setminus \{\phi_i, \phi_j\} \cup \{\sum_W \phi_i * \phi_j\}$.
else
- Set $\Phi_i := \Phi_i \setminus \{\phi_i, \phi_j\} \cup \{\phi_i * \phi_j\}$.

iv. Update $B$ by deleting all pairs $p$ where $\phi_i \in p$ or $\phi_j \in p$.

Until $|\Phi_i| = 1$.

(c) Set $\Phi := \Phi \setminus \Phi^I \cup \Phi_i$.

5. Return $\pi^{\downarrow T} = \pi(\Phi)$.

It is important to temporarily remove *idle* potentials $\Phi^I$ from $\Phi$ in order to maintain the decomposition.

A variable $X$ is eliminated from a combination pair $\{\phi_i, \phi_j\}$ when $\phi_i$ and $\phi_j$ are the only potentials with $X$ in their domains.

**Theorem 7.1 [Lazy SPI Propagation]**
*Suppose we start with a joint potential $\pi_V$ with evidence $\epsilon$ on a junction tree $\mathcal{T}$, and pass messages as described above. When a clique $A$ has received a message from each adjacent clique, the combination of all incoming messages with its own potential is equal to the $A$-marginal of $\pi_V$ w.r.t. $\epsilon$:*

$$\pi_V^{\downarrow A} = (\otimes_{C \in \mathcal{C}} \pi_C)^{\downarrow A} = \pi_A \otimes (\otimes_{C \in \mathrm{adj}(A)} \pi_{C \to A}),$$

*where $\mathcal{C}$ is the set of cliques in $\mathcal{T}$.*

From $\pi_V^{\downarrow A}$, we can compute posterior marginals by marginalization and normalization of $\mathrm{c}(\pi_V^{\downarrow A})$.

## 8 Lazy AR Propagation

AR is a more fine-grained operation than both VE and SPI. The basic idea of AR when computing a single marginal is to perform a sequence of arc-reversals and barren variable eliminations on the DAG of $\mathcal{N}$ until the desired marginal is obtained.

Let $Y$ have parents $\mathrm{pa}(Y) = I \cup J$ and $X$ have parents $\mathrm{pa}(X) = \{Y\} \cup J \cup K$ s.t. $I \cap J = I \cap K = J \cap K = \emptyset$, $I = \mathrm{pa}(Y) \setminus \mathrm{pa}(X)$, $J = \mathrm{pa}(Y) \cap \mathrm{pa}(X)$, and $K = \mathrm{pa}(X) \setminus \mathrm{fa}(Y)$. The reversal of arc $(Y, X)$ proceeds by setting $\mathrm{pa}(Y) = I \cup J \cup K \cup \{X\}$ and $\mathrm{pa}(X) = I \cup J \cup K$ and performing the below computations, see Figure 3:

$$P(X \mid I, J, K) = \sum_Y P(Y \mid I, J) P(X \mid Y, J, K) \quad (2)$$

$$P(Y \mid X, I, J, K) = \frac{P(Y \mid I, J) P(X \mid Y, J, K)}{P(X \mid I, J, K)} \quad (3)$$

Notice, that arc-reversal is not a local computation algorithm in the following sense. When reversing an arc $(Y, X)$, it is necessary to test for existence of a directed path from $Y$ to $X$ not containing $(Y, X)$. If such a path exists, then the arc $(X, Y)$ cannot be reversed before one or more other arcs have been reversed.

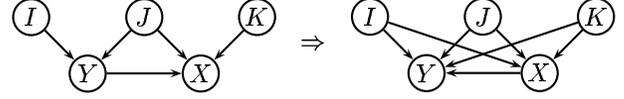

Figure 3: An illustration of Arc-Reversal.

**Algorithm 8.1 [AR Message Computation]**
To solve $Q = (\pi = (\Phi), T, \epsilon)$ by AR do:

1. Moralize $G = G(\pi)$ to obtain $G^m = (V, E)$.

2. Let $\sigma$ be an elimination order for $V \setminus T$.

3. Eliminate each variable $Y \in V \setminus T$ in order of $\sigma$

(a) Set $\Phi_Y := \{\phi \in \Phi \mid Y \in \mathrm{dom}(\phi)\}$.

(b) For each variable $X$ such that $(Y, X) \in E$
i. Let $\phi_Y(Y \mid \mathrm{pa}(Y))$ be the unique potential such that $Y \in \mathrm{H}(\phi)$.
ii. Let $\phi_X(X \mid \mathrm{pa}(X))$ be the unique potential such that $X \in \mathrm{H}(\phi)$.
iii. Reverse $(Y, X) \in E$ by Equation 2 and 3 to obtain $\phi'_X$ and $\phi'_Y$.
iv. Set $\Phi_Y := \Phi_Y \cup \{\phi'_X, \phi'_Y\} \setminus \{\phi_X, \phi_Y\}$.

(c) Set $\Phi := \Phi \cup \Phi_Y \setminus \{\phi(Y \mid \mathrm{pa}(Y))\}$.

4. Return $\pi^{\downarrow T} = \pi(\Phi)$.

In step 3b, we use a topological sort of the original Bayesian network to control the sequence of arc-reversals in order to avoid creating directed cycles.

It is not necessary to perform the last invocation of Equation 3 in step 3(b)iii of the algorithm. When the last arc outgoing from $Y$ say $(Y, X)$ is reversed it is not necessary to compute $\phi'_Y$ as $Y$ will become barren and will be eliminated in step 3c. Notice that the solution of a query will be a set of conditional probability potentials with a single head variable or a single piece of evidence.

**Theorem 8.1 [Lazy AR Propagation]**
*Suppose we start with a joint potential $\pi_V$ with evidence $\epsilon$ on a junction tree $\mathcal{T}$, and pass messages as described above. When a clique $A$ has received a message from each adjacent clique, the combination of all incoming messages with its own potential is equal to the $A$-marginal of $\pi_V$ w.r.t. $\epsilon$:*

$$\pi_V^{\downarrow A} = (\otimes_{C \in \mathcal{C}} \pi_C)^{\downarrow A} = \pi_A \otimes (\otimes_{C \in \mathrm{adj}(A)} \pi_{C \to A}),$$

*where $\mathcal{C}$ is the set of cliques in $\mathcal{T}$.*

Figure 4 illustrates the main motivation for the work presented in this paper. The two possible sequences of arc-reversals leading to the barren variable elimination of $A$ produce two different DAGs $G_{DE}$ (left) and $G_{ED}$ (right). Notice that $G_{DE}$ and $G_{ED}$ capture different conditional independence relations. On the other hand, the result of eliminating $A$ by VE or SPI is a single potential $\phi(D, E \mid B, C)$ representing fewer independence relations. One of the questions considered in this paper is whether or not we can exploit this advantage of AR to improve efficiency of Lazy propagation.

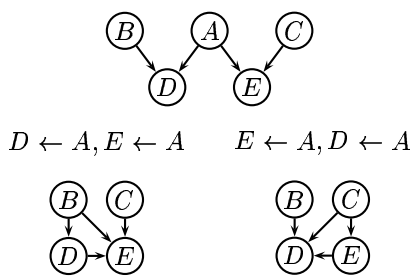

Figure 4: Two different sequences of arc-reversals produce two different DAG structures.

| Network | $\lvert V \rvert$ | $\lvert \mathcal{C} \rvert$ | $max_{C \in \mathcal{C}} s(C)$ | $s(\mathcal{C})$ |
|---|---|---|---|---|
| ship-ship | 50 | 35 | 4,032,000 | 24,258,572 |
| KK | 50 | 38 | 5,806,080 | 14,011,466 |
| net125-7 | 125 | 100 | 37,748,736 | 57,404,777 |
| net150-5 | 150 | 130 | 262,144 | 638,544 |
| net200-5 | 200 | 178 | 15,925,248 | 70,302,065 |

Table 1: Information on the Bayesian networks and their junction trees used in the tests.

## 9 Experiments

In this section, we report on the results of an empirical evaluation of how the performance of lazy propagation depends on the query-based inference algorithm applied to message computation.

We randomly generated a number of test networks with 25 to 200 variables with zero to five parents and two to five states. Ten networks of each size were generated. We report on the results of the empirical evaluation for a subset of these networks and for two real-world Bayesian networks. Table 1 describes the complexity of the selected networks and their junction trees. In the table, $s(C) = \prod_{X \in C} \lVert X \rVert$ is the state space size of clique $C$ where $\lVert X \rVert$ denotes the state space size of $X$, and $s(\mathcal{C}) = \sum_{C \in \mathcal{C}} s(C)$ is the total clique state space size where $\mathcal{C}$ is the set of cliques of the junction tree.

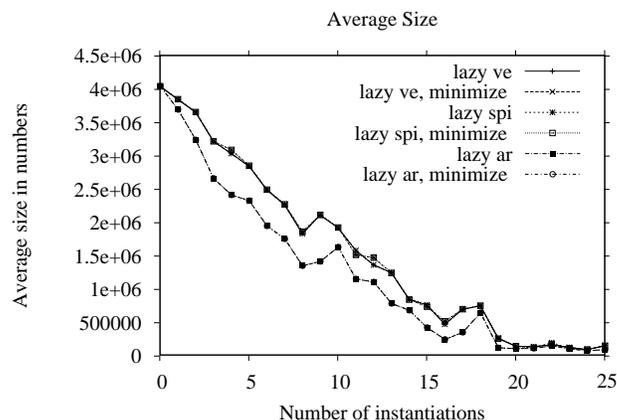

Figure 5: Largest potential size for ship-ship.

For each network, the size of the evidence set varied from 0 to 25 instantiated variables. For each size of the evidence set, we randomly generated 25 sets of evidence. Figures 5 - 7 show the average size of the largest potential created during inference, while Figures 9 and 10 show the average time for inference in seconds. The state space size $s(\phi)$ of a potential $\phi$ is defined as $s(\phi) = \prod_{X \in \text{dom}(\phi)} \lVert X \rVert$.

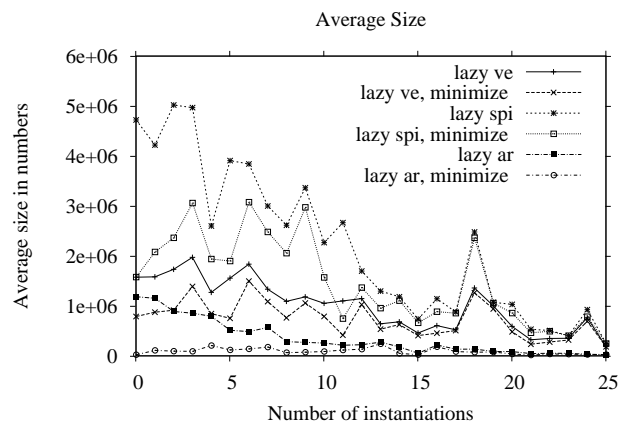

Figure 6: Largest potential size for net125-7.

For Lazy AR propagation and Lazy VE propagation, on-line triangulation was performed using a heuristic method with one-step lookahead. The minimum-fill-in-weight heuristic was chosen based on our own experiments and the experiments of (Kjærulff 1993). For Lazy SPI propagation, we performed an implicit on-line triangulation using the heuristic method with one-step lookahead as suggested by (Li and D'Ambrosio 1994). Off-line triangulations were determined based on a graph decomposition by minimal separators approach, see (Jensen 2004) and citations therein. The optimality criteria used is total clique weight (i.e., to-tal clique state space size).

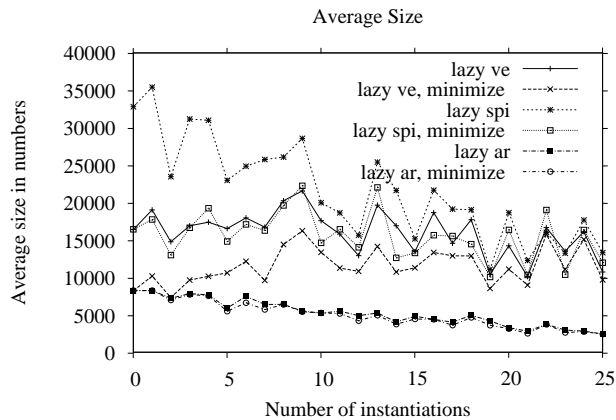

Figure 7: Largest potential size for net150-5.

The results show that in most cases the performance of Lazy propagation for probabilistic inference was rather insensitive to the choice of VE, SPI, or AR as the message computation algorithm. In some cases AR did produce better performance of inference w.r.t. both time and space complexity. Figure 5 shows that for the *ship-ship* network AR produced better results than VE and SPI for the space costs of inference. This improved efficiency in space came at a cost of an increase in time cost. The figure also show that minimalization of potential tails did not have any impact. For other networks minimalization did not reduce the size of the largest potential, but nevertheless gave an improvement in time efficiency.

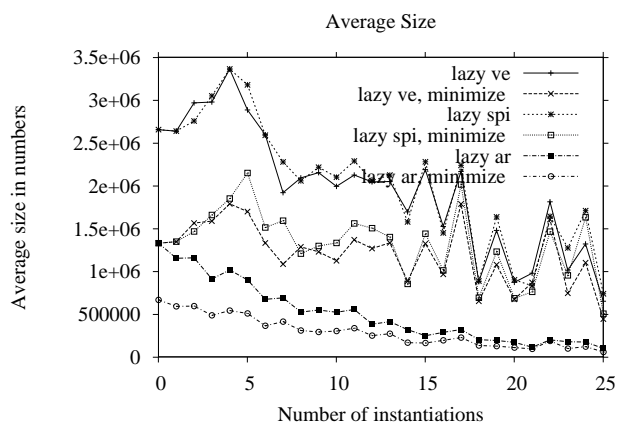

Figure 8: Largest potential size for net200-5.

Figure 6 shows an example where minimalization of potential tails had an impact on the space cost of inference whereas Figure 7 shows an example where minimalization had an impact for both SPI and VE, but not for AR. Figures 8 and 9 show that minimalization may produce a reduction in both time and space costs of inference. Also on this network, AR produced the best results. Figure 10 shows an example where minimalization of tails induce a higher time cost of inference with AR being slightly faster than SPI and VE except for zero and one instantiated variables.

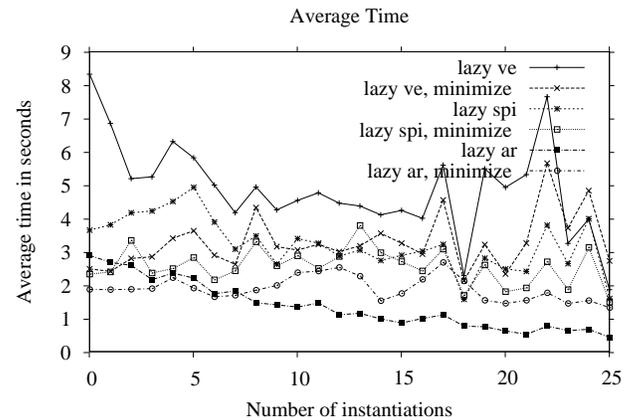

Figure 9: Time for net200-5.

The difference in efficiency of SPI and VE is most likely due to different heuristic methods for determining the potential combination and variable elimination orders, respectively. This is a topic of future work.

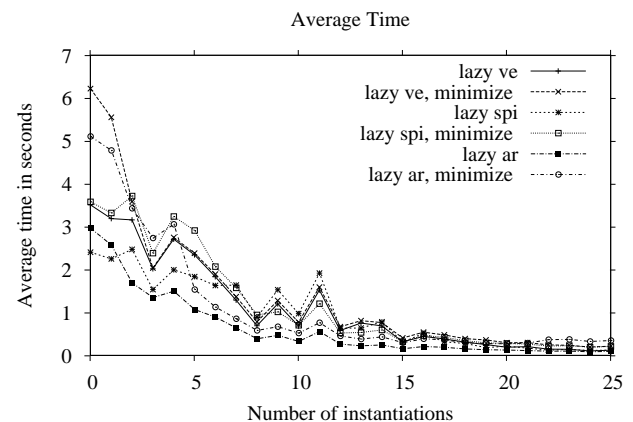

Figure 10: Time for KK.

The experiments were performed using a Java implementation running on a PC with a 2.2 GHz AMD Athlon$^{TM}$ CPU and 768 MB RAM running Redhat 8.

## 10 Discussion and Conclusion

In this paper, we have showed how AR and SPI both may be used as the message computation algorithm of Lazy propagation. The results of the empirical evalu-ation show that SPI and AR may equally well be used as the message computation algorithm. For some networks, Lazy AR propagation may even offer better performance than both Lazy VE and Lazy SPI propagation. Notice that the average size of the largest potential is almost always significantly smaller than the largest clique in the junction tree. Both space and time cost of inference are reduced with the number of instantiated variables in all three architectures.

The experiments indicate that under some circumstances Lazy AR propagation is able to exploit the properties of barren variables and independence relation induced by evidence better than Lazy SPI/AR propagation. This is due to the fact that the Lazy AR propagation algorithm maintains a decomposition of clique and separator potentials were each factor is a potential with a single head variable. This more fine-grained decomposition in some cases allows for the exploitation of additional independence and irrelevance properties during inference. Notice that we have used the same heuristic triangulation method for Lazy AR and Lazy VE. Hence, the difference in performance must be due to maintaining an orientation of all edges in Lazy AR.

We also investigated the impact of potential tail minimalization on the efficiency of inference. Minimalization is implemented by performing a separation analysis on the graph of the Bayesian network for each tail variable in each potential of each message. This implies a relatively large computational overhead, which is only justified when it produces smaller potentials. The results show that in some cases minimalization of tails produced a smaller average largest potential at a cost of an increase in inference time. Minimalization in some cases led to a reduction in both time and space costs (figures 8 and 9).

The empirical results presented in this paper have encouraged the idea of applying Lazy AR Propagation to other types of probabilistic graphical models such as mixed Bayesian networks and mixed influence diagrams where the work of (Shachter and Kenley 1989; Poland 1994) on solving mixed models using arc-reversal can be exploited. Furthermore, we would like to investigate the performance of Lazy AR propagation in combination with algorithms for exploiting independence of causal influence.